\documentclass{llncs}
\pdfoutput=1
\usepackage{graphicx}
\begin{document}
\setlength{\parindent}{12pt}
\title{Dynamic Risk Assessment for Vehicles of Higher Automation Levels by Deep Learning}
\author{Patrik Feth\inst{1}, Mohammed Naveed Akram\inst{1}, Ren{\'e} Schuster\inst{2}, Oliver Wasenm{\"u}ller\inst{2}}
\institute{Fraunhofer Institute for Experimental Software Engineering
\email{name.surname@iese.fraunhofer.de}
\and DFKI - German Research Center for Artificial Intelligence
\newline
\email{name.surname@dfki.de}
}
\maketitle	
\begin{abstract}
Vehicles of higher automation levels require the creation of situation awareness. One important aspect of this situation awareness is an understanding of the current risk of a driving situation. In this work, we present a novel approach for the dynamic risk assessment of driving situations based on images of a front stereo camera using deep learning. To this end, we trained a deep neural network with recorded monocular images, disparity maps and a risk metric for diverse traffic scenes. Our approach can be used to create the aforementioned situation awareness of vehicles of higher automation levels and can serve as a heterogeneous channel to systems based on radar or lidar sensors that are used traditionally for the calculation of risk metrics.  
\end{abstract}
\section{Introduction}
\label{introduction_section}
The contribution of this paper is a novel method for the creation of situation awareness regarding risk, i.e. \textbf{dynamic risk assessment}, from monocular images and disparity maps created from a stereo camera. Stereo cameras are widely used in cars nowadays and can be considered as standard \cite{schuster2018sceneflowfields}. We trained a Convolutional Neural Network (CNN) with supervised learning to derive the risk of a driving scene from an image of that particular scene. To achieve this, we first created a data set of driving situations in a simulation environment and annotated those situations with a risk metric. This data was then given as the input to the learning process. After the training, the CNN was able to predict the current risk of a driving situation from a camera input. This functionality can be used as a part of vehicles of higher automation levels. \\ \indent
In general, there are two main approaches followed for achieving higher automation levels: The robotics approach \cite{Thrun.2006} and the deep learning approach \cite{Bojarski.2016}. The robotics approach is more traditional and considers the vehicle as a combination of different functions realized by sensors, control logic and actuators. This approach is the one followed mainly by the traditional car industry. The deep learning approach is very different from that. It considers the vehicle as a black box into which sensor information is fed, this information is processed by a neural network and control output as steering and throttle commands are generated by the network. This approach is the one mainly followed by new entries to the automotive market. The neural network can be trained with millions of recorded driving kilometers. The feasibility of the approach was shown in 1989 by a team from CMU \cite{Pomerleau.1989} and was repeated by a team from NVIDIA using the advances of machine learning in 2016 \cite{Bojarski.2016}. Even though this approach is feasible, a huge and open question arises towards its dependability. It is currently a hot topic to try to understand what a neural network has learned and why decisions are made the one way or the other \cite{Bojarski.2016.2}. However, reliable results on those questions can probably not be   expected by the time that car manufacturers want to have realized highly automated vehicles.\\ \indent
The approach presented in this paper can be used in both paradigms for the development of vehicles of higher automation levels presented above. The main and currently considered use case lies in the robotic case, where it can be used as an additional component of the functional architecture for performing a dynamic risk assessment. A further and future use case of our approach lies in the field of making the output of a neural network more understandable and thus analyzable. If the deep learning paradigm is used for achieving higher automation levels, the risk shall become an additional output of the neural network that controls the vehicle. This allows making the output of the neural network more transparent, as additional information about the network's current understanding of the environment is available. This understanding can then be plausibilized and it can be checked if the predicted control commands of the neural network match the predicted risk of the current driving scene.\\ \indent
This paper is structured as follows: In Section \ref{relatedWork_section} we compare the presented approach to the state-of-the-art, which is - as of today - only a single comparable approach for the dynamic risk assessment based on images. Section \ref{solution_section} gives details on our approach, the data collection and the training process. Section \ref{results_section} gives the results of the training process and the achieved performance of our approach. Section \ref{conclusion_section} concludes the paper.

\section{Related Work}
\label{relatedWork_section}
The approach related closest to ours, is the work presented in \cite{Wang.2017}. The goal of this approach is - just as in ours - to train a Convolutional Neural Network to perform dynamic risk assessment of traffic scenes from images of that particular scene.\\ \indent 
The authors of \cite{Wang.2017} used a subjective classification for the rating of traffic scenes. They showed YouTube videos from collisions recorded by dashboard cameras to three different subjects. These subjects then had to assign two time labels for each video: $t_1$ as the time at which the root cause for the accident appears and $t_2$ as the time at which this cause becomes obvious. This divides the video into three sections: a \textit{normal}, a \textit{caution} and a \textit{warning} section. The CNN in their work was trained to assign one of those three labels to a given image of a traffic scene. In our work we used a more objective measure to classify traffic scenes by the use of a quantitative and objective risk metric, as will be explained in Section \ref{metric_section}. By that, our network is solving a regression instead of a classification problem. Conclusively, the approach presented in our work is more systematic in creating the ground truth data for classifying traffic scenes. \\ \indent
In addition to the work in \cite{Wang.2017} more traditional approaches for risk assessment that work with explicit trajectory predictions and the calculation of conflicts between trajectories can be considered as related work. A representative of those approaches is described in \cite{Schreier.2014}. The solution in this paper differs from the traditional approaches for risk assessment as it follows an end-2-end paradigm deriving the current risk directly from the camera input. Such an end-2-end paradigm can result in a significant saving of computational costs. However, the state of research is still too early to evaluate any claims of such kind.

\section{Data Generation and Training of the CNN}
\label{solution_section}
\subsection{Overview}
\label{solution_overview_section}
\begin{figure}
\centering
\includegraphics[width=1\textwidth]{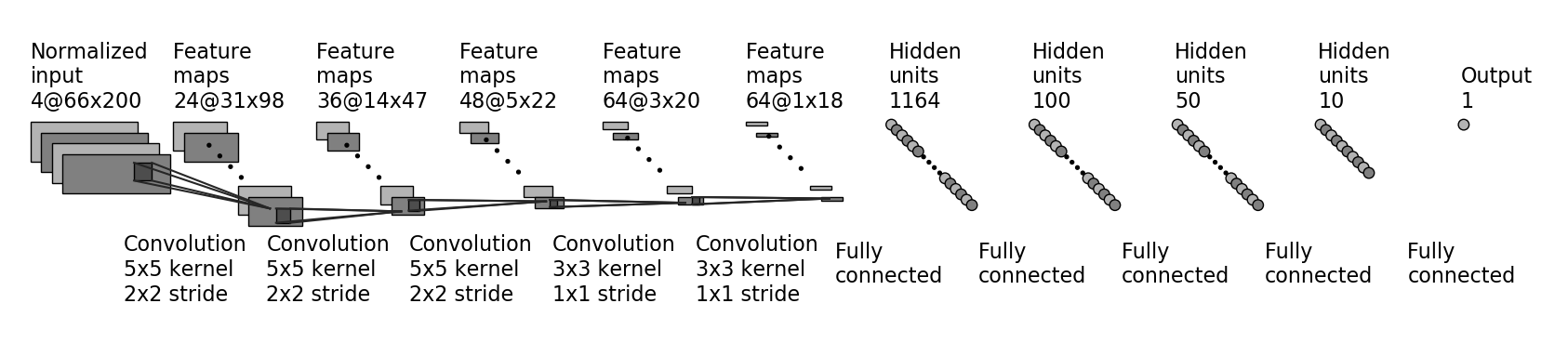}
\caption{The architecture of our neural network for risk estimation in driving scenarios} 
\label{architecture_figure}
\end{figure}
The recent success of neural networks in almost all fields of Computer Vision has motivated us to address the problem of dynamic risk assessment with a deep Convolutional Neural Network (CNN).
Convolutional Neural Networks are a special type of Artificial Neural Networks (ANNs). While general ANNs model a function as connected neural nodes of different layers, in CNNs these connections are structured as convolutional filter kernels \cite{lecun2015deep}. That means that the connections between layers are very dense whereas the number of parameters is kept relatively small because the connections share filter weights in a patch-wise manner. In addition, modeling a neural network by multiple convolutions is intuitively related to workflows of classical image processing. For example, some kernels of the first two convolutional layers in a CNN can often be interpreted as first and second order derivative filters \cite{zeiler2014visualizing}, which are also commonly used in classical edge detection. As framework for the design and training of our neural network, we used TensorFlow (https://www.tensorflow.org/). \\ \indent
Since visual sensor data is available in most vehicles nowadays, we decided to design the network in an end-to-end manner, i.e. the input is image data and the output is the final risk estimate (cf. Figure \ref{architecture_figure}). As others before, we wanted to predict the output based on images from an onboard camera of a moving vehicle. Consequently, our network architecture is inspired by the existing DAVE2 network \cite{Bojarski.2016} that is predicting steering commands for highly automated vehicles. \\ \indent
A full overview of the design is given in Figure \ref{architecture_figure}. In detail, the network consists of five convolutional layers and five fully connected layers. All layers except the last one use Rectified Linear Units (ReLU) as activation function. Since we used  subsampled resolution of the input images, we did not need any spatial pooling. As mentioned earlier, we provided a single image and a precomputed disparity map as input for the network and stacked them into a single four channel tensor comparable to RGB-D images. More details on how these data was generated is given in Section \ref{data_section} and Section \ref{metric_section}. The first three convolutional layers use kernels of size $5\times 5$ with a stride of $2\times 2$. The last two convolutional layers apply $3\times 3$ kernels densely, i.e. the stride is 1. The first fully connected layer after the flattened final convolution has 1.164 output nodes. All following fully connected layers steadily reduce the number of neurons to a single output node.
\subsection{Simulation Environment}
To train a CNN for dynamic risk assessment with supervised learning, a great amount of training data was required. It is challenging to create this amount of data in a real driving environment. Thus, we used a simulation environment to create sufficient training data. This environment needs to be large and random enough to be representative for a real environment. For research purposes, it is common to use computer games as a substitute for very expensive professional simulation tools. In the past, efforts have been made to use the TORCS game \cite{DeepDrivingTORCS.2015} for training neural networks to drive a car autonomously. This simulation environment is open source but lacks a sufficient graphical appearance and is limited to simulate driving situation on a race track. A more suited candidate for a simulation environment was Grand Theft Auto - V. With more than 500 vehicle models, urban and countryside road networks, and random actions of pedestrians and other drivers, it comes close to an ideal simulation environment for autonomous vehicles. Although the game is developed for commercial purposes, it’s publisher Rockstar Games has allowed creative, non-commercial modification of the game under certain conditions (https://support.rockstargames.com/hc/en-us/articles/200153756-Policy-on-posting-copyrighted-Rockstar-Games-material). Existing work already shows that data from this particular gaming environment is valid data for the benchmarking of object classification algorithms \cite{Richter.2017}. Visual surveillance is another application of commercial games as simulation environment for a serious purpose \cite{OVVV.2007}. During creating this work, an open-source simulator for autonomous driving research named CARLA \cite{Dosovitskiy17} was released. For future work we plan on using this environment as it has almost the same benefits as GTA V only with a smaller simulated world and it was designed for research purposes, which means that the interface is easier to handle and the license is more open. 
\subsection{Creating Stereo-Images and Disparity}
\label{data_section}
\begin{figure}
\centering
\includegraphics[width=1\textwidth]{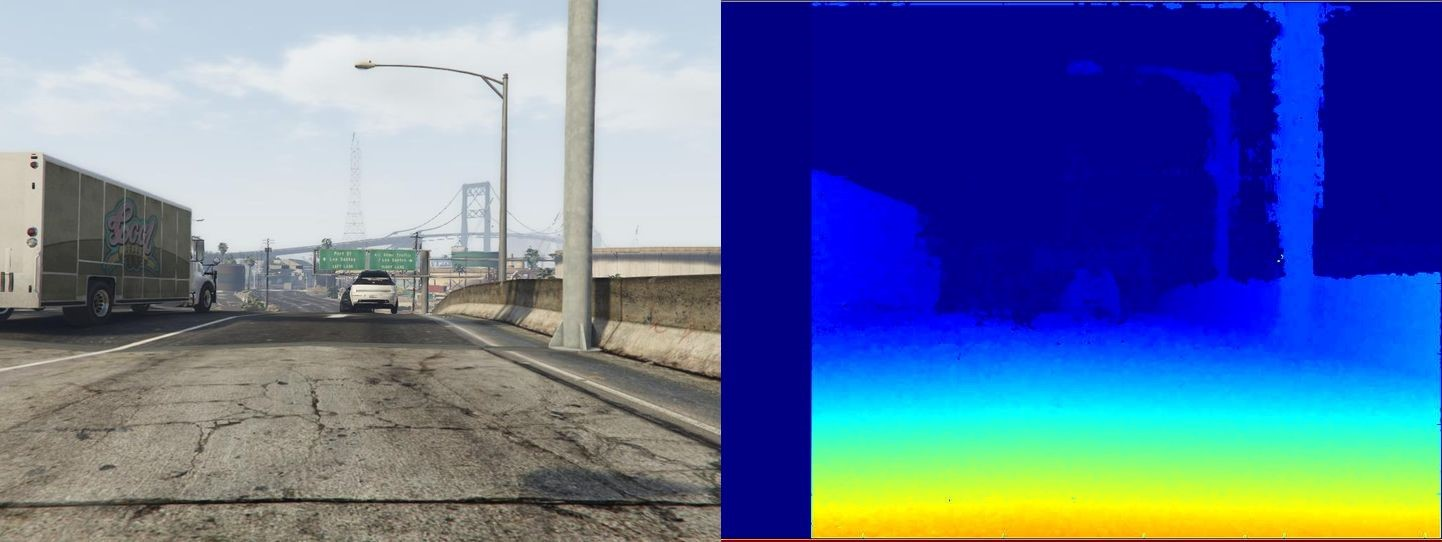}
\caption{Sample disparity map} 
\label{Disparity}
\end{figure}

In order to determine the risk of a given situation from an image, the distance of the objects - like other cars, pedestrians, etc. - in the scene to the vehicle is essential. From a monocular image, it is very difficult to estimate these distances, since no 3D information is available. Cheap active sensors - such as depth cameras - have difficulty to convince in automotive applications \cite{yoshida2017time}. Thus, we used a stereo camera, which has a left and a right camera. By computing the shift in pixels between these two cameras, we got a disparity image. Disparity is inversely proportional to the distance from the camera to the respective object. \\ \indent
We rendered the two images in such a way that both images lie on the same plane and are horizontally aligned. With that, the corresponding image points in the two images share the same horizontal line, which reduces the search space from two to one dimension. For the computation of disparity several algorithm exist. We choose the state-of-the-art Semi Global Matching (SGM) algorithm \cite{hirschmuller2008SGM} for our approach. SGM tries to find the best correspondences for all pixels of one image so that a consistent and smooth disparity image is created. It utilizes a semi-global energy formulation using eight different paths radiating from each target pixel location. This enables sharp boundaries, accurate disparity estimation and fast run time. Because of occlusions, it is hard to recover disparity for all image points. SGM uses a left-right consistency check for two computed disparity images to localize and remove wrong values. Thus, SGM yields a non-dense disparity image. \\ \indent
We created stereo images out of GTA V by simulating a standard stereo camera. We virtually placed two custom cameras on the hood of a car with the same orientation and same Y and Z-axis position. We kept the X-axis, i.e. the horizontal position, 16cm apart, simulating the baseline of a stereo camera. For consistency, we used the same car model for our data extraction process. We rendered these custom cameras and automatically took screenshots to obtain left and right images for computing disparity. GTA V allows only to render one camera per frame. Thus, we first rendered the left camera, slowed down the game to the lowest possible timescale, and then rendered the right camera in the next frame. Hence, there was one frame delay in the two images. This delay of one frame translated to a timely delay of 88 milliseconds. We used feature matching to verify if the difference between the two images is as intended. This was performed by matching ORB features \cite{ORB.2011} in both images and computing the average difference in Y (vertical) and X (horizontal) direction of those identified features for each image pair. Ideally, only a distance in X direction should exist. The difference in Y direction should be zero. A difference in Y direction can only happen because of the time delay between the images or the inaccuracies of ORB matching. We classified image pairs with an average X difference less than 1 pixel, or average Y difference greater than 5 pixels as outliers. In the created data set of 110.152 images around 3.5 \% outliers were found and excluded. An example of a disparity map created from a traffic scene is shown in the right part of Figure \ref{Disparity}. In this visualization of the disparity map yellow corresponds to a close object and blue corresponds to an object farer away. 
\\
\\
\subsection{Risk Metric}
\label{metric_section}
\begin{figure}
\includegraphics[width=\textwidth]{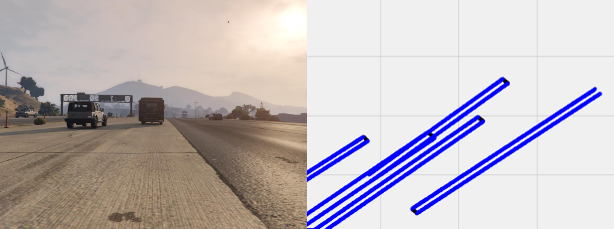} 
\caption{Left: Scene from the simulation environment | Right: Corresponding grid map}
\label{grid_example}
\end{figure}

\begin{figure}
\includegraphics[width=\textwidth]{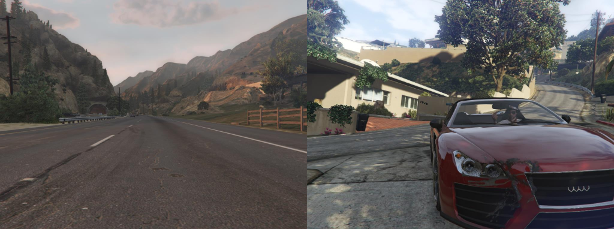} 
\caption{Left: Uncritical scene (0) | Right: Critical scene (0.21)}
\label{rmc_comparison}
\end{figure}

The recorded scenarios from the simulation environment were imported into a Risk Metric Calculator (RMC) to determine the ground truth values for the dynamic risk assessment. Risk metrics are used extensively  for the creation of situation awareness of active safety systems in the automotive domain. The ISO standard for forward collision avoidance systems \cite{ISO.2013} uses thresholds of the metrics \textit{Time To Collision (TTC)} and \textit{Enhanced Time To Collision (ETTC)} to decide if and if so which collision mitigation strategy shall be activated. Those two metrics work with a very simple constant turn assumption for the lateral movement and constant velocity (for TTC)/constant acceleration (for ETTC) assumptions for the longitudinal movement. There exist more mature risk metrics that use richer prediction models for the dynamic risk assessment, for example \cite{Schreier.2014} or \cite{Berthelot.2012}. The availability of such a great set of metrics with different limitations and assumptions makes the choice of the right metric a difficult task. Feth et al. \cite{Feth.2017} presents strategies on how to come to a valid decision regarding this aspect.\\ \indent
To get a deeper understanding and the possibility to compare the values of different risk metrics in traffic situations, we used a Risk Metric Calculator (RMC). More detailed information on the RMC tool is available in \cite{Feth.tobepublished}. The RMC takes driving scenarios, either generated in the tool itself or given as a record from the simulation environment, fills for every simulation step a grid map with the relevant information and uses that grid map to calculate the selected risk metrics. Grid maps are a common notation in the robotics world. They discretize the environment in cells of fixed or variable sizes. Each cell contains information about its occupancy status. If it is occupied it may contain information about the object that is occupying the cell. This map represents the World Model for the robot, e.g. the vehicle, and is used to determine the appropriate behavior, e.g. the desired trajectory. See \cite{Jungnickel.2016} for related work on the subject and a more recent approach for the creation of a grid map. In Figure \ref{grid_example} we illustrate how an exemplary grid map for a specific driving scene in our approach looked like. This grid map shows the cells that are currently occupied by the subject vehicle (in the center of the map) and the vehicles in the vicinity in black color and those cells that will be occupied in the next three seconds using a constant turn and constant speed assumption in blue color. For efficiency reasons the occupied cells are approximated by the "border cells". Cells within these borders are not filled to keep the amount of entries in the grid map small. The horizontal and vertical lines in the visualization have no meaning.\\ \indent
\\
For the development of our new approach, we decided to first use a very simple risk metric. We investigated if we can train a Convolutional Neural Network to predict this simple risk metric. The reason for choosing such a simple risk metric first was the required amount of data. Usually, the risk metrics with more sophisticated prediction models have a lower false positive rate. That means that for most traffic scenes they do not indicate any risk. Consequently, to achieve the required amount of data we would have needed to increase drastically the amount of training images in the simulation environment. The risk metric used in this work is the reciprocal of the \textit{time headway} metric.
\begin{equation}
time\_headway = \frac{distance}{subject\_velocity}
\end{equation} 
The RMC calculated this metric by counting the least number of empty cells between one vehicle and another vehicle and dividing that distance by the current speed of the subject vehicle. By using the reciprocal of that value, we got a positive relationship between the value of the metric and the current risk: the higher the value of the metric, the higher the risk of the current driving situation. Only vehicles in front of the subject vehicle were considered. If there were multiple other vehicles in the vicinity of the subject vehicle, the RMC considered the closest vehicle only. The result was afterwards limited to a value of 20 to exclude outliers and then normalized to a range between 0 and 1. See Figure \ref{rmc_comparison} for an example of a driving situation that achieved a low value for the specified risk metric (0) and another driving situation that achieved a high value for the specified risk metric (0.21). A value of 0.21 means that following a constant speed and pessimistic turn assumption for the subject vehicle and assuming that the target vehicle will not move at all, a crash would happen in 0.238 seconds. The \textit{pessimistic turn assumption} used here leads to a \textit{worst time headway} metric comparable to the \textit{worst time to collision} metric presented in \cite{Wachenfeld.2016}. The actual trajectory of neither the subject nor the target vehicle are considered in the calculation of the metric. Again, the motivation for using this risk metric, which obviously leads to a overestimation of the risk of driving situations, was the needed effort in the production of a sufficient large set of training data and the fact that this work shall be a proof of concept.\\ \indent
\\ 
Finally, this ground truth data was used as input for training the CNN. The trained CNN is the result of this paper and our novel solution for dynamic risk assessment for vehicles of higher automation levels by deep learning.  Instead of calculating the risk metric for a recorded driving scenario, this CNN can be used together with a camera to perform a dynamic risk assessment of the driving situation while driving.
\subsection{Network Training}
We used the TensorFlow framework for training. We built the network architecture as discussed in Section \ref{solution_overview_section} and trained the network using Adam optimizer on the collected data. A total of 110.152 data points were recorded. After estimating the quality using ORB as described in Section \ref{data_section}, we excluded outliers and used the remaining 106.170 data points effectively in the training. This data set consisted of the left image, the disparity information calculated from the stereo image and the calculated risk value from the RMC tool. The network was given the left monocular image and the disparity information as input and was trained to predict the risk value as output. We divided the full data set into three random groups. 70 \% of the data was used for training, 10 \%  for validation and 20 \% was used for final testing. Training was performed for 30 epochs with a batch size of 10. To avoid overfitting, validation test was performed after every 10 steps.
\section{Results}
\label{results_section}
\begin{figure}
\centering
\includegraphics[width=0.8\textwidth]{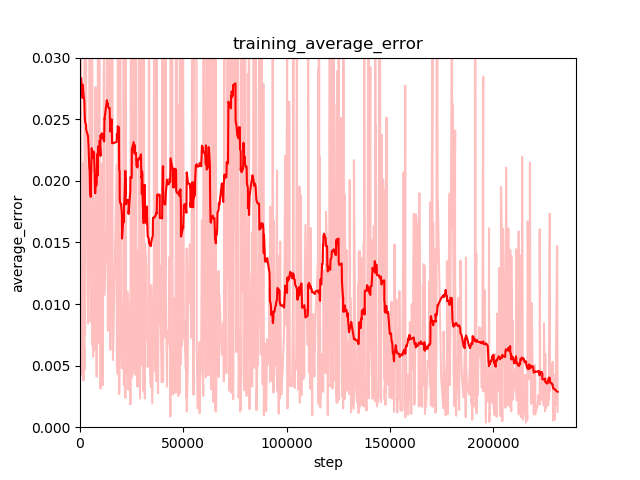}
\caption{Training Average Error} 
\label{DispMonoTrainingError}
\end{figure}

\begin{figure}
\centering
\includegraphics[width=0.8\textwidth]{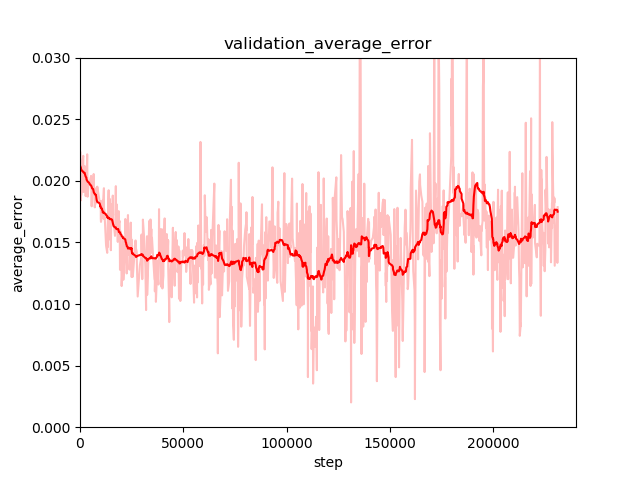}
\caption{Validation Average Error} 
\label{DispMonoValidationError}
\end{figure}

\begin{figure}
\includegraphics[width=\textwidth]{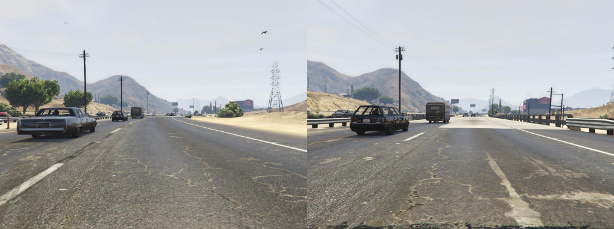} 
\caption{Left: Value from RMC: 0.36 Value from CNN: 0.38 | Right: Value from RMC: 0.75 Value from CNN: 0.39}
\label{nn_performance}
\end{figure}

The model was trained to minimize the mean squared error between true and predicted risk values on the training data set. The mean squared error on the training data set is referred to as \textit{training average error}. Validation was performed on the validation dataset and hence, its error is referred to as \textit{validation average error}. Likewise, testing error is referred to as \textit{testing average error}. The final \textit{testing average error} we achieved over all test data was \textbf{0.01470}. The values for \textit{training average error} and \textit{validation average error} are shown in Figures \ref{DispMonoTrainingError} and \ref{DispMonoValidationError}. \\ \indent
It can be seen from the curve of the training error in Figure \ref{DispMonoTrainingError}, that the network was able to learn an estimation of risk from the provided input data. The training error was steadily decreasing. However, a common pitfall for all machine learning methods is overfitting. That means that the network loses the ability to generalize to unknown input because it starts to create too strong connections between training input and output. Therefore, we monitored the validation error in Figure  \ref{DispMonoValidationError} on a separate set of data. After half the iterations, i.e. after approximately 15 epochs, the validation error started to increase. That is the optimal point to stop training. As usual, overall performance of the network can be increased by providing more, distinctive training data, though this implies more training effort. \\ \indent
Figure \ref{nn_performance} gives an exemplary impression of the performance of our CNN by showing a driving situation for which the risk metric was predicted correctly by the network (RMC value 0.36, network value 0.38) and a driving situation for which the risk metric was not predicted correctly by the network (RMC value 0.75, network value 0.39). \\ \indent
For a better interpretation of the performance of the CNN we have to consider the application in which it is used. A system for dynamic risk assessment needs to classify the current situation as either critical or uncritical. The performance of the network shall thus be rated according to its ability to correctly classify a driving situation. Assuming a time headway value of 1.5 seconds as the threshold, the complete effective data set of 106.170 situations contains 76.789 critical driving situations. Of those 76.789 situations the CNN classified 67.052 correctly. Table \ref{confusion_matrix} gives the complete confusion matrix. With those values our approach reached an \textbf{accuracy of 72.87 \%}. From a functional safety perspective false positives (FP) and false negatives (FN) are most relevant. The acceptable FP and FN ratio are determined by the risk that follows from a wrong classification. This depends on the application in which the dynamic risk assessment functionality is used. If the result of the dynamic risk assessment is used to trigger a very critical driving maneuver, as for example an emergency brake, in an automated driving system, the acceptable false positive rate is very low. If the result is used to only trigger a visual driver warning, false negatives might be more critical but still not too critical. Defining the target ratios is within the scope of functional safety engineering and shall not be covered here. However, the values that we achieve for our solution show that the integrity of the developed approach is still far from what can be considered as usable in a safety-critical context.\\ \indent
\begin{table}
\centering
\begin{tabular}{|l|c|c|c|}
	\hline 
	& Actual Critical Situation & Actual Uncritical Situation  \\ 
	\hline 
	Predicted Critical Situation & 67052 (63,16 \%) & 19069 (17,96 \%) \\ 
	\hline 
	Predicted Uncritical Situation & 9737 (9,17 \%) & 10312 (9,71 \%) \\ 
	\hline 
\end{tabular} 
\caption{Confusion Matrix of the CNN}
\label{confusion_matrix}
\end{table}
\\
We could identify a major reason for the bad performance of the network: The view of the camera did not perfectly match the area that the RMC considered in the calculation. Because of the that, the most critical situations according to the RMC happened outside the view of the camera and could therefore not be derived from the image. The example in Figure \ref{nn_performance} illustrates this: On the left image our CNN was most probably considering the vehicle on the left as the closest vehicle. In the following, this vehicle was passed and on the right image, it is right next to the subject vehicle but not visible for the camera. In the RMC tool, this occluded vehicle was used for the calculation of the risk metric. However, the network was considering the vehicle in front. This vehicle was just as far away as the vehicle on the image before, giving a reason why the same value was predicted by the network as for the scene before. \\ \indent
With an accuracy value of 72.87 \% our CNN does not perform significantly better than the related work in \cite{Wang.2017}. The authors of \cite{Wang.2017} achieved a maximum accuracy of 70.77 \%. However, the classification of traffic situations in our work is based on an objective risk metric instead of a subjective labeling.
\section{Conclusion}
\label{conclusion_section}
In this paper, we presented a novel approach for the dynamic risk assessment from images. To achieve this, we trained a Convolutional Neural Network with data generated in a simulation environment and annotated with a simple risk metric calculated in a risk metric calculation tool. We extended the simulation environment with the capability to generate stereo images. From those stereo images, disparity maps have been generated. We annotated monocular images and the disparity maps with the reciprocal of the \textit{time headway} risk metric. The trained network achieved an accuracy of \textbf{72.87 \%} on our test set. \\ \indent 
In the introduction of this paper we already pointed at the reliability challenge of neural networks. Of course, the same holds for the CNN used in this paper. Because of that, a system as recommended in this paper can only be used in addition to more conservative systems for dynamic risk assessment. As we transition with higher automation levels from fail safe to fail operational systems, measures for redundancy will be required. The CNN proposed in this paper can be such a redundant system with a high degree of heterogeneity from conservative systems working with radar and lidar information. In parallel to this work we currently investigate means to safely integrate systems for dynamic risk assessment as presented in this work in the architecture of safety-critical systems.\\ \indent
For future work, we plan to go from the simple \textit{time headway} risk metric to a more sophisticated risk metric with a richer collision prediction model. This will require enhancing the training data, as situations rated as critical will be less likely. Further, we plan to use different network structures, as e.g. Long / Short Term Memory (LSTM) Networks and compare the performance to the base performance achieved in this paper. Once we achieve a sufficient performance, the approach should be transferred from the simulation environment to a real environment.

\bibliographystyle{splncs04}
\bibliography{bibliography}
\end{document}